\documentclass[10pt,twocolumn,letterpaper]{article}

\usepackage{cvpr}
\usepackage{times}
\usepackage{epsfig}
\usepackage{graphicx}
\usepackage{amsmath}
\usepackage{amssymb}
\usepackage{multirow}
\usepackage{balance}

\usepackage{soul}
\setul{}{1pt}

\usepackage[pagebackref=true,breaklinks=true,letterpaper=true,colorlinks,bookmarks=false]{hyperref}

\cvprfinalcopy 

\def\cvprPaperID{4277} 

\ifcvprfinal\fi

\begin{document}

\title{Video Relationship Reasoning using Gated Spatio-Temporal Energy Graph}

\author{Yao-Hung Hubert Tsai$^\dagger$, Santosh Divvala$^\ddagger$, Louis-Philippe Morency$^\dagger$, Ruslan Salakhutdinov$^\dagger$, Ali Farhadi$^{\ddagger *}$\\
$^\dagger$Carnegie Mellon University, $^\ddagger$Allen Institute for AI, $^*$University of Washington\\
{\color{red} \url{ https://github.com/yaohungt/GSTEG_CVPR_2019}}
}

\maketitle

\begin{abstract}
\vspace{-2mm}

   Visual relationship reasoning is a crucial yet challenging task for understanding rich interactions across visual concepts. For example, a relationship \{man, open, door\} involves a complex relation \{open\} between concrete entities \{man, door\}. While much of the existing work has studied this problem in the context of still images, understanding visual relationships in videos has received limited attention. Due to their temporal nature, videos enable us to model and reason about a more comprehensive set of visual relationships, such as those requiring multiple (temporal) observations (e.g., \{man, lift up, box\} vs. \{man, put down, box\}), as well as relationships that are often correlated through time (e.g., \{woman, pay, money\} followed by \{woman, buy, coffee\}). In this paper, we construct a Conditional Random Field on a fully-connected spatio-temporal graph that exploits the statistical dependency between relational entities spatially and temporally. We introduce a novel gated energy function parametrization that learns adaptive relations conditioned on visual observations. Our model optimization is computationally efficient, and its space computation complexity is significantly amortized through our proposed parameterization. Experimental results on benchmark video datasets (ImageNet Video and Charades) demonstrate state-of-the-art performance across three standard relationship reasoning tasks: Detection, Tagging, and Recognition.
\end{abstract}


\vspace{-5mm}
\section{Introduction}


Relationship reasoning is a challenging task that not only involves detecting low-level entities (subjects, objects, etc.) but also recognizing the high-level interaction between them (actions, sizes, parts, etc.). Successfully reasoning about relationships not only enables us to build richer question-answering models (e.g., {\em Which objects are larger than a car?}), but also helps in improving image retrieval~\cite{lu2016visual}(e.g., images with {\em elephants drawing a cart}), scene graph parsing~\cite{zellers2018neural} (e.g., {\em woman has helmet}), captioning~\cite{zhang2017visual}, and many other visual reasoning tasks. 


\begin{figure}[t!]
\begin{center}
\includegraphics[width=0.43\textwidth]{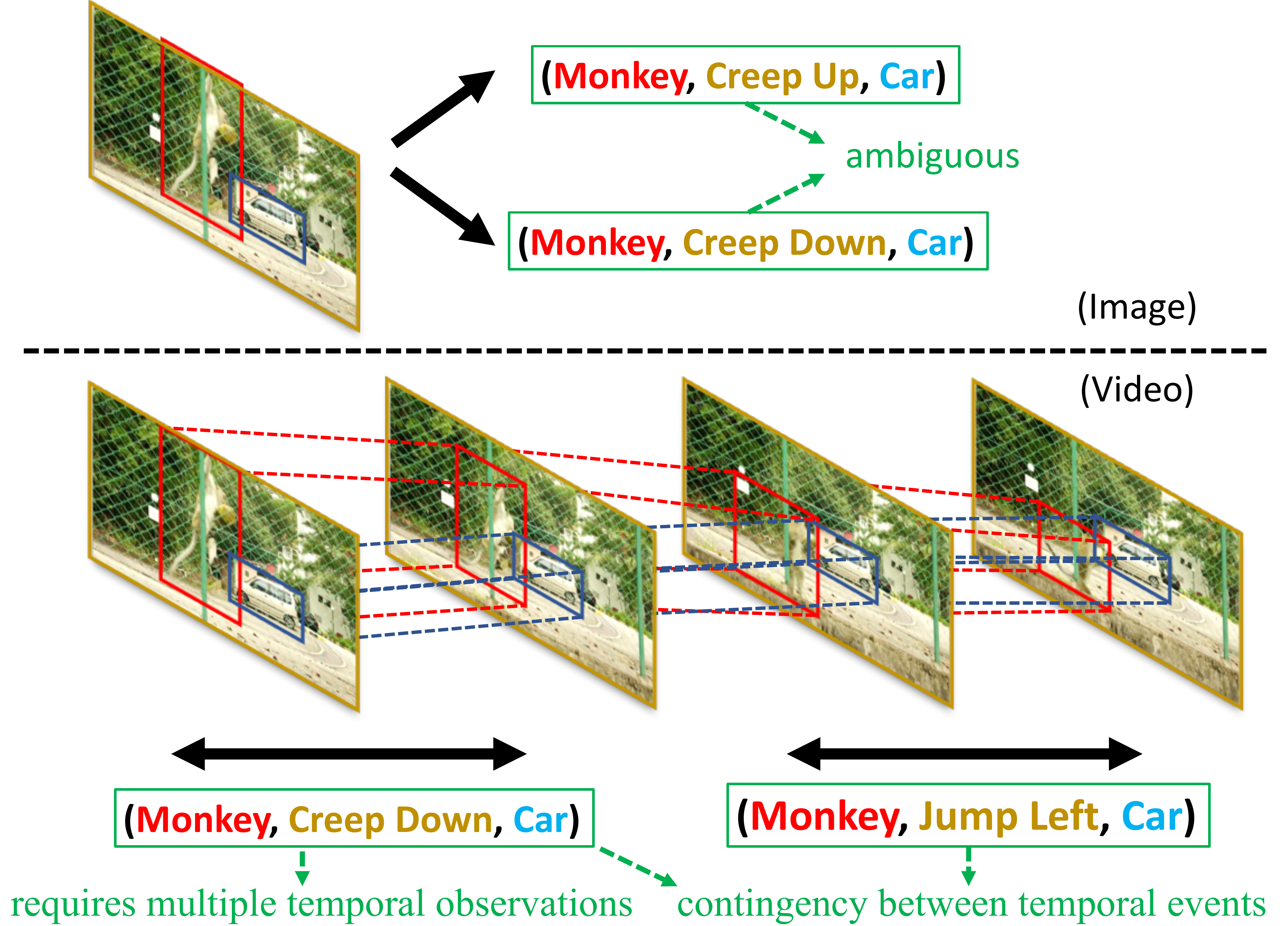}
\end{center}
\vspace{-4mm}
\caption{\small Visual relationship reasoning in images (top) vs. videos (bottom): Given a single image, it is ambiguous whether the {\em monkey} is creeping up or down the {\em car}. Using a video not only helps to unambiguously recognize a richer set of relations, but also model temporal correlations across them (e.g., {\em creep down} and {\em jump left}).}
\label{fig:illus1}
\vspace{-3mm}
\end{figure}

\begin{figure*}[t!]
\vspace{-3mm}
\includegraphics[width=\textwidth]{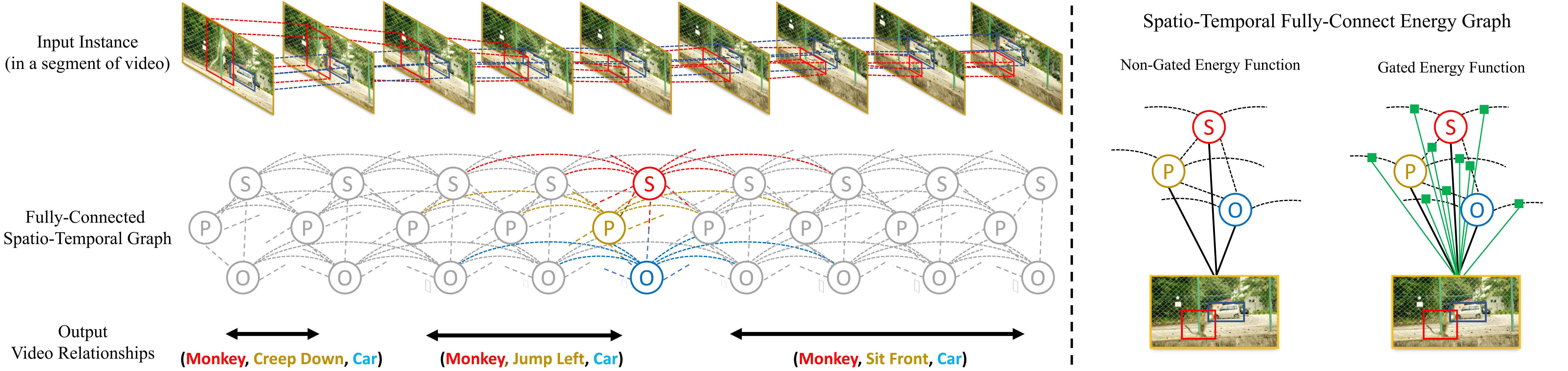}
\vspace{-4mm}
\caption{An overview of our Proposed Gated Spatio-Temporal Energy Graph. Given an input instance (a video clip), we predict the output relationships (e.g., \{{\em monkey, creep down, car\}}, etc.,) by reasoning over a fully-connected spatio-temporal graph with nodes $\mathbf{S}$ (Subject), $\mathbf{P}$ (Predicate) and $\mathbf{O}$ (Object). Unlike previous works that assumed a non-gated (i.e., predefined or globally-learned) pairwise energy function, we explore the use of gated energy functions (i.e., conditioned on the specific visual observation) . Best viewed zoomed in and in color.}
\label{fig:illus2}
\vspace{-5mm}
\end{figure*}

Most contemporary research in visual relationship reasoning has been focused in the domain of static images. While this has resulted in several exciting and attractive reasoning modules~\cite{sadeghi2011recognition,lu2016visual,zhang2017visual,liang2017deep,yin2018zoom,zhu2018deep,cui2018context,liang2018visual}, it lacks the ability from reasoning about complex relations that are inherently temporal and/or correlated in nature. For example, in Fig.~\ref{fig:illus1} it is ambiguous to infer from a static image whether the monkey is creeping down or up the car. Also, it is difficult to model relations that are often correlated through time, such as {\em man enters room} and {\em man open door}.

In this paper, we present a novel approach for reasoning about visual relationships in videos. Our proposed approach jointly models the spatial and temporal structure of relationships in videos by constructing a fully-connected spatio-temporal graph (see Fig.~\ref{fig:illus2}). We refer to our model as a Gated Spatio-Temporal Energy Graph. In our graph, each node represents an entity and the edges between them denote the statistical relations. Unlike much of the previous work~\cite{krahenbuhl2011efficient,zheng2015conditional,schwing2015fully,dai2017detecting,sigurdsson2017asynchronous} that assumed a predefined or globally-learned pairwise energy function, we introduce an observation-gated version that allows us to make the statistical dependency between entities adaptive (conditioned on the observation). 

Our adaptive parameterization of energy function helps us model the natural diversification of relationships in videos. For instance, the dependency between {\em man} and {\em cooking} should be different conditioned on the observation (i.e., whether the location is {\em kitchen} or {\em gym}). However, given the large state space of observations (in videos), directly maintaining observation-dependent statistical dependencies may be computationally intractable~\cite{mnih2013playing,tsai2017discovering}. Towards this end, we develop an amortized parameterization of our new gated pairwise energy function, which combines ideas from clique template~\cite{taskar2002discriminative,taylor2009factored,mccallum2009factorie}, neural networks~\cite{goodfellow2016deep,tsai2017discovering}, and tensor factorization~\cite{koren2009matrix} for achieving efficient inference and learning. 

We evaluate our model on two benchmark datasets, ImageNet Video~\cite{ILSVRC15} and Charades~\cite{sigurdsson2016hollywood}. Our method achieves state-of-the-art performance across three standard relationship reasoning tasks: detection, tagging, and recognition. We also study the utility of our model in the zero-shot setting and learning from semantic priors.

\section{Related Work}

\noindent {\bf Video Activity Recognition.} The notion of activity in a video represents the interaction between objects~\cite{goyal2017something,kay2017kinetics} or the interaction between an object and a scene~\cite{sigurdsson2016hollywood}. While related to our task of relation reasoning, activity recognition does not require explicit prediction of all entities, such as subject, object, scene, and their relationships. The term {\em relation} used in activity recognition and relationship reasoning has different connotations. In the visual relationship reasoning literature, it refers to the correlation between different entities, such as object, verb, and scene, while in activity recognition, it refers to either correlation between activity predictions (i.e., single entity) or correlation between video segments. For example, \cite{zhou2017temporal} proposed Temporal Relation Network to reason the temporal `relations' across frames at multiple time scales. \cite{girdhar2017actionvlad} introduced a spatio-temporal aggregation on local convolutional features for better learning representations in the video. \cite{wang2018non} proposed Non-Local Neural Networks to model pairwise relations for every pixel in the feature space from low-layers to higher-layers. The work was extended to~\cite{wang2018videos} for constructing a Graph Convolutional Layer that further modeled relation between object-level features. 


\noindent {\bf Visual Relationship Reasoning.} Most recent works in relation reasoning have focused their analysis on static images~\cite{yin2018zoom,zhu2018deep,cui2018context,liang2018visual}. For example, \cite{sadeghi2011recognition} introduced the idea of visual phrases for compositing visual concepts of subject, predicate, and object. \cite{lu2016visual} decomposed the direct visual phrase detection task into individual detection on subject, predicate, and object leading to improved performance. \cite{dai2017detecting} further applied conditional random fields on top of the individual predictions to leverage their statistical correlations. \cite{liang2017deep} proposed a deep variation-structured reinforcement learning framework and then formed a directed semantic action graph. The global interdependency in this graph facilitated predictions in local regions of the image. One of the key challenges of learning relationships in videos has been the lack of relevant annotated datasets. In this context, the recent work of~\cite{shang2017video} is inspiring as it contributes manually annotated relations for the ImageNet video dataset. Our work improves upon ~\cite{shang2017video} on multiple fronts: (1) Instead of assuming no temporal contingency between relationships, we introduce a gated fully-connected spatio-temporal energy graph for modeling the inherently rich structure from videos; (2) We extend the study of relation triplet from subject/predicate/object to a more general setting, such as object/verb/scene~\cite{sigurdsson2016hollywood}; (3) We consider a new task `relation recognition' (apart from relation detection and tagging) which requires the model to make predictions in a fine-grained manner; (4) For various metrics and tasks, our model demonstrates improved performance.

\noindent {\bf Deep Conditional Random Fields.} Conditional Random Fields (CRFs) have been popularly used to model the statistical dependencies among predictions in images~\cite{he2004multiscale,zheng2015conditional,schwing2015fully,sadeghi2015viske,dai2017detecting} and videos~\cite{quattoni2007hidden,sigurdsson2017asynchronous}. Several extensions have been recently introduced for fully-connected CRF graphs. For example,~\cite{zheng2015conditional,schwing2015fully,sigurdsson2017asynchronous} attempted to express fully-connected CRFs as recurrent neural networks and made the whole network end-to-end trainable, which has led to interesting applications in image segmentation~\cite{zheng2015conditional,schwing2015fully} and video activity recognition tasks~\cite{sigurdsson2017asynchronous}. In the characterization of CRFs, the unary energy function represents the inverse likelihood for assigning a label, while the binary energy function measures the cost of assigning multiple labels jointly. However, most of the existing parameterizations of binary energy functions~\cite{krahenbuhl2011efficient,zheng2015conditional,schwing2015fully,dai2017detecting,sigurdsson2017asynchronous} have limited or no connections to observed variables. Such parameterizations may not be optimal for video relationship reasoning due to the adaptive idiosyncrasy for statistical dependencies between entities. To address the issue, we instead propose an observation-gated pairwise energy function with efficient and amortized parameterization.

\vspace{-1mm}
\section{Proposed Approach}
\label{sec:prop}

The task of video relationship reasoning not only requires modeling the entity predictions spatially and temporally, but also maintaining a changeable correlation structure between entities across videos with various contents. To this end, we propose a Gated Spatio-Temporal Fully-Connected Energy Graph for capturing the inherently rich video structure into account. 

We start by defining our notations using Fig.~\ref{fig:illus2} as a running example. The input instance $X$ lies in a video segment and consists of $K$ synchronous input streams $X = \{X^k\}_{k=1}^K$. In this example, input streams are \{object trajectories, predicate trajectories, subject trajectories\}, and thus $K=3$, where trajectories refer to the consecutive frames or bounding boxes in the video segment. Each input stream contains observations for $T$ time steps (i.e., $X^k = \{X^k_t\}_{t=1}^{T}$), where for example object trajectories represent object bounding boxes through time. For each input stream, our goal is to predict a sequence of entities (labels) $Y^k = \{Y^k_t\}_{t=1}^T$. In Fig.~\ref{fig:illus2}, the output sequence of predicate trajectories represent predicate labels through time. Hence we formulate the data-entities tuple as $(X, Y)$ with $Y = \{Y^1_t, Y^2_t \cdots, Y^K_t\}_{t=1}^{T}$ representing a set of sequence of entities. 

The entity $Y^k_t$ should spatially relate to entities $\{\{Y^1_t, Y^2_t \cdots, Y^K_t\} \setminus \{Y^k_t\}\}$ and temporally relate to entities $\{\{Y^k_1, Y^k_2 \cdots, Y^k_T\} \setminus \{Y^k_t\}\}$. For example, suppose that the visual relationships observed in a grocery store are \{\{mother, pay, money\}, \{infant, get, milk\}, \{infant, drink, milk\}\}; spatial correlation must exist between mother/pay/money and temporal correlation must exist between pay/get/drink. We also note that implicit correlation may also exist between $Y^k_t$ and $Y^{k'}_{t'}$ for $t \neq t', k \neq k'$. 
Based on the structural dependencies between entities, we propose to construct a Spatio-Temporal Fully-Connected Energy Graph (see Sec.~\ref{subsec:struc_pred}), where each node represents an entity and each edge denotes the statistical dependencies between the connected nodes. To further take account that the statistical dependency between ``get'' and ``drink'' may be different depending on different observations (i.e., location in grocery store v.s. home), we introduce an observation-gated parameterization for pairwise energy functions. In the new parameterization, we amortize the potentially large computational cost by using clique templates~\cite{taskar2002discriminative,taylor2009factored,mccallum2009factorie}, neural network approximation~\cite{mnih2013playing,tsai2017discovering}, and tensor factorization~\cite{koren2009matrix} (see Sec.~\ref{subsec:amadmessage}). 


\subsection{Spatio-Temporal Fully-Connected Graph}
\label{subsec:struc_pred}
By treating the predictions of entities as random variables, the construction of the graph can be realized by forming a Markov Random Field (MRF) conditioned on a global observation, which is the input instance (i.e., $X$). Then, the tuple ($X, Y$) can be modeled as a Conditional Random Field (CRF) parametrized by a Gibbs distribution of the form: $
P\Big(Y = y|X\Big) = \frac{1}{Z(X)}\mathrm{exp}\Big(-\mathbf{E}(y|X)\Big),
$
where $Z(X)$ is the partition function and $\mathbf{E}(y|X)$ is the energy of assigning labels $Y =y = \{y^1_t, y^2_t, \cdots, y^K_t\}_{t=1}^T$ conditioned on $X$. Assuming only pairwise cliques in the graph \big(i.e., $P(y|X):= P_{\psi, \varphi}(y|X), \mathbf{E}(y|X) := \mathbf{E}_{\psi,\varphi}(y|X)$\big), the energy can be expressed as:
\vspace{-1mm}
\begin{equation}
\small
\mathbf{E}_{\psi,\varphi}(y|X)  
= \sum_{t,k} \psi_{t,k}(y_t^k|X) + \sum_{\{t,k\}\neq \{t',k'\}} \varphi_{t,k,t',k'}(y_t^k, y_{t'}^{k'}|X),
\label{eq:energy}
\end{equation}
where $\psi_{t,k}$ and $\varphi_{t,k,t',k'}$ are the unary and pairwise energy, respectively. In Eq.~\eqref{eq:energy}, the unary energy, which is defined on each node in the graph, captures inverse likelihood for assigning $Y_t^k = y_t^k$ conditioned on the observation $X$. Typically, this term can be derived from an arbitrary classifier or regressor, such as a deep neural network~\cite{lecun2015deep}. On the other hand, the pairwise energy models interactions of label assignments across nodes $Y_t^k = y_t^k, Y_{t'}^{k'} = y_{t'}^{k'}$ conditioned on the observation $X$. Therefore, the pairwise term determines the statistical dependencies between entities spatially and temporally. However, the parameterization in most previous works on fully-connected CRF~\cite{zheng2015conditional,schwing2015fully,sigurdsson2017asynchronous,dai2017detecting} assumes that the pairwise energy function is non-adaptive to the current observation, which may not be ideal to model changeable dependencies between entities across videos. In the following Sec.\ref{subsec:amadmessage}, we propose an observation-gated parametrization for pairwise energy function to address the issue. 
\subsection{Gated Pairwise Energy Function}
\label{subsec:amadmessage}

Much of existing work uses a simplified parameterization of pairwise energy function and typically considers only the {\em smoothness} of the joint label assignment. For instance, in Asynchronous Temporal Field~\cite{sigurdsson2017asynchronous}, $\varphi_\cdot(y_t^k, y_{t'}^{k'}|X)$ is defined as $\mu(y_t^k, y_{t'}^{k'})K(t, t')$, where $\mu$ represents the label compatibility matrix and $K(t,t')$ is an affinity kernel measurement which represents the discounting factor between $t$ and $t'$. Similarly, in the image segmentation domain~\cite{zheng2015conditional,schwing2015fully}, $\varphi_\cdot(s_i, s_j|I)$ is defined as $\mu(s_i, s_j)K(I_i, I_j)$, where $s_{\{i,j\}}$ is the segment label and $I_{\{i,j\}}$ is the input feature for location $\{i,j\}$ in image $I$. In these models, the pairwise energy comprises an observation-independent label compatibility matrix followed by a spatio or temporal discounting factor. We argue that the parametrization of pairwise energy function should be more expressive.
To this end, we define the pairwise energy as:
\begin{equation}
\begin{split}
\varphi_{t,k,t',k'}(y_t^k, y_{t'}^{k'}|X) & := \left \langle f^\varphi \right \rangle_{X, t, t',k,k',y_t^k, y_{t'}^{k'}},
\end{split}
\end{equation}
where $f^\varphi$ can be seen as a discrete lookup table that takes the input $X$ of size $|X|$ and outputs a large transition matrix of size $(T^2K^2-1)\times |Y_t^k|\times |Y_{t'}^{k'}|$, and where $\left \langle \cdot \right \rangle_z$ represents its $z_{th}$ item. Directly maintaining this lookup table is computationally intractable due to the large state space of $X$. Considering a simple case that $X$ is a pairwise-valued $32\times 32$ image, we have $|X| = 2^{32\times 32}$ possible states. The state space complexity aggravates when $X$ becomes an RGB-valued video. Thanks to the recent advances in graphical models~\cite{taskar2002discriminative,taylor2009factored,mccallum2009factorie}, deep neural networks~\cite{mnih2013playing,tsai2017discovering}, and tensor factorization~\cite{koren2009matrix}, our workaround is to parametrize and approximate $f^\varphi$ as $f^\varphi_\theta$ with learnable parameters $\theta$ as follows:
\begin{equation}
\begin{split}
&\left \langle f^\varphi \right \rangle_{X, t, t',k,k',y_t^k, y_{t'}^{k'}} \approx f^\varphi_\theta (X_t^k,t,t',k,k',y_t^k, y_{t'}^{k'}) \\
= & \left\{\begin{matrix}
 \left \langle g^{kk'}_\theta(X_t^k)  \otimes h^{kk'}_\theta(X_t^k)  \right \rangle_{y_t^k, y_{t'}^{k'}}  & t = t' \\ K_\sigma \Big(t,t'\Big) 
 \left \langle r^{kk'}_\theta(X_t^k)  \otimes s^{kk'}_\theta(X_t^k)\right \rangle_{y_t^k,y_{t'}^{k'}}  & t\neq t'
\end{matrix}\right. ,
\end{split}
\label{eq:pair}
\end{equation}
where $g^{kk'}_\theta(\cdot), r^{kk'}_\theta(\cdot) \in \mathbb{R}^{|Y_t^k|\times r}$ and $h^{kk'}_\theta(\cdot),s^{kk'}_\theta(\cdot)  \in \mathbb{R}^{|Y_{t'}^{k'}|\times r}$ represent the $r$-rank projection from $X_t^k$, which is modeled by a deep neural network. $A \otimes B = AB^\top$ denotes the function on matrix $A$ and $B$, and results in a transition matrix of size $|Y_t^k|\times |Y_{t'}^{k'}|$.  $K_\sigma \Big(t,t'\Big)$ is the Gaussian kernel with bandwidth $\sigma$ representing discounting factor for different time steps.

The intuition behind our parametrization is as follows: First, we note that clique templates~\cite{taskar2002discriminative,taylor2009factored,mccallum2009factorie} are adopted spatially and temporally, which leads to scalable learning and inference. Second, the idea of using neural networks for approximating the lookup tables ensures both parameter efficiency and generalization~\cite{goodfellow2016deep,tsai2017discovering}. The lookup table maintains the state transitions of $\mathcal{X} \rightarrow \mathcal{Y}^k \times \mathcal{Y}^{k'}$ where calligraphy font denotes the corresponding state space. Finally, we choose $r << \mathrm{min}_{k}\{|Y_t^k|\}$ so that a low-rank decomposition is performed on the transition matrix from $Y_t^k$ to $Y_{t'}^{k'}$. 
The low-rank decomposition allows us to substantially reduce the number of learnable parameters.   
To summarize, our design for $f^\varphi_\theta$ amortize the large space complexity for $f^\varphi$ and is gated by observation.

 \subsection{Inference, Message Passing, and Learning}
\label{subsec:infer}

Minimizing the CRF energy in Eq.~\eqref{eq:energy} returns the most probable label assignment problem of $Y=\{y^1_t, y^2_t, \cdots, y^K_t\}_{t=1}^T$ given the observation $X$. However, the exact inference in a fully connected CRF is often computationally intractable even with variables enumeration or elimination~\cite{koller2009probabilistic}. 
In this work, we adopt the commonly used mean-field algorithm~\cite{koller2009probabilistic} as approximate inference, which finds the approximate posterior distribution $Q(Y)$ such that $Q(\cdot)$ is closest to $P_{\psi, \varphi}(Y|X)$ in terms of $\mathcal{KL}(Q//P_{\psi, \varphi})$ within the class of distributions representable as a product of independent marginals $Q(Y) = \prod_{t,k} Q(Y_t^k)$.
Following~\cite{koller2009probabilistic}, inference can now be realized as the naive mean-field updates with the coordinate descent optimization, and it can be expressed in terms of fixed-point message passing equations:
\vspace{-1mm}
\begin{equation}
Q(y_t^k) \propto \Psi_{t,k}\Big({y_t^k|X}\Big) \prod_{\{t', k'\} \neq \{t, k\}}m_{t',k',t,k}(y_{t}^{k}|X)
\label{eq:variational}
\end{equation}
\vspace{-1mm}
\\with $\Psi_{t,k} = \mathrm{exp}\Big(-\psi_{t,k}\Big)$ representing the unary potential and $m_\cdot(\cdot)$ denoting the message having form\footnote{In Supplementary, we make connection from our gated amortized parametrization for pairwise energy function in message form with Self-Attention~\cite{vaswani2017attention} in machine translation and Non-Local Means~\cite{buades2005non} in Image Denoising.} of
\begin{equation}
m_{\cdot}(\cdot) =\mathrm{exp}\Big( - \sum_{y_{t'}^{k'}}\varphi_{t,k,t',k'}(y_t^k, y_{t'}^{k'}|X)Q(y_{t'}^{k'}) \Big).
\label{eq:message}
\end{equation}

To parametrize the unary energy function, we use a similar formulation:
\vspace{-2mm}
\begin{equation}
\begin{split}
\psi_{t,k} (y_t^k|X) := & \left \langle f^\psi \right \rangle_{X , t, k, y_t^k} \\
 \approx & f^\psi_\theta(X_t^k, t, k, y_t^k) = \left \langle w^k_\theta (X_t^k) \right \rangle_{y_t^k},
\end{split}
\label{eq:unary}
\end{equation}
\vspace{-3mm}
\\where $w_\theta^k \in \mathbb{R}^{|Y_t^k|}$ represents the projection from $X_t^k$ to logits of size $|Y_t^k|$, modeled by a deep neural network.

Lastly, we cast the learning problem as minimizing conditional cross-entropy between the proposed distribution and the true one, where $\theta$ denotes the parameters we need in our model: $\theta^* = \mathrm{arg\,min}_\theta \,\,\mathbb{E}_{X, Y}[-\mathrm{log}\,Q (Y)]$.

\begin{table*}[t!]
\begin{center}
\fontsize{7pt}{10pt}
\selectfont
\vspace{-4mm}
\begin{tabular}{|cc||ccc||ccc||cccc|}
\hline
\multirow{3}{*}{Method} & Correponding & \multicolumn{3}{c||}{Relationship Detection}       & \multicolumn{3}{c||}{Relationship Tagging}            & \multicolumn{4}{c|}{Relationship Recognition}                           \\ &
                        Image-Relationship or                 & \multicolumn{3}{c||}{relationship}         & \multicolumn{3}{c||}{relationship}            & subject        & predicate      & object         & relationship \\
                                &    Video-Activity Method      & R@50          & R@100         & mAP           & P@1            & P@5            & P@10           & Acc@1          & Acc@1          & Acc@1          & Acc@1            
                \\ \hline \hline 
\multicolumn{12}{|c|}{Standard Evaluation} \\ \hline \hline
VidVRD$^*$~\cite{shang2017video}                          &    Visual Phrases~\cite{sadeghi2011recognition}    & 5.58          & 6.68          & 6.94          & 41.00          & 29.60          & 21.85          & 80.28          & 16.55          & 80.40          & 12.93            \\
UEG                         &  VRD$_V$~\cite{lu2016visual}    & 2.81        & 3.64          & 2.94      & 31.50          & 19.88          & 14.98           & 80.15          & 23.95          & 80.55          & 18.62            \\
UEG$^\dagger$  & VRD~\cite{lu2016visual}                                &        3.41       &     4.05 &      4.52   &   36.00         &     21.60        & 15.41           &   80.15     & 25.92     & 80.55       & 22.47               \\
SEG            &  DRN~\cite{dai2017detecting}        &   4.34         &      5.32      & 4.16       &     35.00         &    27.10          &    20.90         & 85.15          & 25.85          & 84.26          & 20.97            \\
STEG           &     AsyncTF~\cite{sigurdsson2017asynchronous}        &    4.18        &   4.98          &    4.71        &     40.00      &   24.45       &  17.66            & 89.91         & 25.92          & 89.33          & 22.54            \\ 
\textbf{GSTEG (Ours)}       &     -       & \textbf{7.05} & \textbf{8.67} & \textbf{9.52} & \textbf{51.50} & \textbf{39.50} & \textbf{28.23} & \textbf{90.60} & \textbf{28.78} & \textbf{89.79} & \textbf{25.01}   \\ \hline \hline 
\multicolumn{12}{|c|}{Zero-Shot Evaluation}   \\ \hline \hline
VidVRD$^*$~\cite{shang2017video}     &         Visual Phrases~\cite{sadeghi2011recognition}                & 0.93        & 1.16       & \textbf{0.18}          & 0.0       & 0.82         & 0.82          & 74.54   & 2.78         & 74.07         & 1.62            \\
UEG               &    VRD$_V$~\cite{lu2016visual}     & 0.0     & 0.23       & 1.30$\times$1e-5         & 0.0       & 0.27        & 0.82       & 74.31   & 5.09    & 74.77     & 3.24           \\
UEG$^\dagger$               &    VRD~\cite{lu2016visual}                                 & 0.23 & 0.23 & 5.36$\times$1e-5 & 0.0 & 0.82 & 0.82 & 78.24 & 5.79 & 78.47 & 3.47                        \\
SEG        &         DRN~\cite{dai2017detecting}                 &   0.23       &  0.46            & 6.70$\times$1e-5       &    0.0       &   0.82 &  1.23            & 81.02   & 6.94        & 74.54        & 3.47          \\
STEG     &         AsyncTF~\cite{sigurdsson2017asynchronous}        &   0.23         &    0.69           &   0.02           &  1.37             &   1.10             &  0.96           & 80.09         & 7.18        & 79.17         & 4.40          \\ 
\textbf{GSTEG (Ours)}         &      -    & \textbf{1.16} & \textbf{2.08} & 0.15 & \textbf{2.74} & \textbf{1.92} & \textbf{1.92} & \textbf{82.18} & \textbf{7.87} & \textbf{79.40} & \textbf{6.02}   \\ \hline
\end{tabular}
\end{center}
\vspace{-2mm}
\caption{Evaluation for different methods on ImageNet Video dataset. $^*$ denotes the re-implementation of~\cite{shang2017video} after fixing the bugs in their released method code (by contacting authors). $^\dagger$ denotes the implementation with additional triplet loss term for language priors~\cite{lu2016visual}. }
\label{tbl:imagenet}
\vspace{-4mm}
\end{table*}

\section{Experimental Results \& Analysis}

In this section, we report our quantitative and qualitative analyses for validating the benefit of our proposed method. 
Our experiments are designed to compare different baselines and ablations for detecting and tagging relationships given a video as well as recognizing relationships in a fine-grained manner. \\



\noindent {\bf Datasets.} We perform our analysis on two datasets: ImageNet Video~\cite{ILSVRC15} and Charades~\cite{sigurdsson2016hollywood}. (a) {\em ImageNet Video}~\cite{ILSVRC15} contains videos (from daily-life as well as in-the-wild) with manually labeled bounding boxes for objects. We utilize the annotations from~\cite{shang2017video}, in which a subset of the videos having rich visual relationships were selected ($1,000$ videos in total with $800$ for training \& rest for evaluation, available at~\cite{VidVRDURL}). The visual relationship is defined on the triplet \{{\em subject}, {\em predicate}, {\em object}\}. For example, \{{\em person}, {\em ride}, {\em bicycle}\} or \{{\em dog}, {\em larger}, {\em monkey}\}, etc. It has $35$ categories of {\em subject} and {\em object}, and $132$ categories of {\em predicate} (see Suppl. for details) with trajectory denoting consecutive bounding boxes through time. A relation triplet is labeled on a pair of trajectories (one for subject and another for object). The entire video has multiple pairs of trajectories and these pairs may or may not overlap with each other spatially or temporally. 
(b) {\em Charades}~\cite{sigurdsson2016hollywood} contains videos of human indoor activities ($9,848$ in total with $7,985$ for training and the rest for evaluation, available at~\cite{CharadesURL}) . The visual relationship is defined on the triplet \{{\em verb}, {\em object}, {\em scene}\} or \{{\em object}, {\em verb}, {\em scene}\}. For example, \{{\em hold}, {\em blanket}, {\em bedroom}\}, \{{\em someone}, {\em cook}, {\em kitchen}\}, etc. It has $33$ categories of {\em verb}, $38$ {\em objects} and $16$ {\em scenes} (see Suppl. for details). Different from ImageNet Videos, as suggested by~\cite{sigurdsson2016hollywood,sigurdsson2017asynchronous,wang2018non,wang2018videos}, we treat the entire video as an input instance. Therefore, a video comprises multiple relation triplets, and each relation triplet is defined within a time segment. The relation triplets may or may not overlap temporally with each other. \\ 

\begin{table*}[t!]
\begin{center}
\footnotesize
\fontsize{7pt}{10pt}
\selectfont
\vspace{-4mm}
\begin{tabular}{|cc||ccc||ccc||cccc|}
\hline
\multirow{3}{*}{Method} & Correponding & \multicolumn{3}{c||}{Relationship Detection}       & \multicolumn{3}{c||}{Relationship Tagging}            & \multicolumn{4}{c|}{Relationship Recognition}                           \\ &
                        Image-Relationship or                  & \multicolumn{3}{c||}{relationship}         & \multicolumn{3}{c||}{relationship}            & object        & verb      & scene         & relationship \\
                                &   Video-Activity Method       & R@50          & R@100         & mAP           & P@1            & P@5            & P@10           & Acc@1          & Acc@1          & Acc@1          & Acc@1            
                \\ \hline \hline
VidVRD~\cite{shang2017video}                         &    Visual Phrases~\cite{sadeghi2011recognition}    & 13.62          &    18.36           &     3.12     &    3.97         &      4.62         &   4.26         & 28.70      & 63.64    & 34.91         & 7.83             \\
UEG                        &  VRD$_V$~\cite{lu2016visual}    & 22.53         & 29.70         & 7.93       & 16.05       & 11.47         & 8.72           & 41.74    & 64.70         & 34.62         & 11.94            \\
UEG$^\dagger$                         &  VRD~\cite{lu2016visual}    & 22.35         & 29.65         & 7.90       & 16.10       & 11.38         & 8.67           & 41.70    & 64.73         & 35.17         & 11.85            \\
SEG            &  DRN~\cite{dai2017detecting}        &  23.68          &    31.56           &     8.77     &    18.04       &      12.50         &   9.37         & 42.84      & 64.36    & 35.28         & 12.60          \\
STEG           &     AsyncTF~\cite{sigurdsson2017asynchronous}         &    23.79            &      31.65        &  8.84           &   18.46          &    12.57         & 9.37            & 42.87  & 64.53   & 35.71        & 12.76               \\ 
\textbf{GSTEG (Ours)}       &     -       & \textbf{24.95} & \textbf{33.37} & \textbf{9.86} & \textbf{19.16} & \textbf{12.93} & \textbf{9.55} & \textbf{43.53} & \textbf{64.82} & \textbf{40.11} & \textbf{14.73}    \\  \hline
\end{tabular}
\end{center}
\vspace{-2mm}
\caption{Evaluation for different methods on Charades dataset. Our method outperforms all competing baselines across the three tasks. }
\label{tbl:charades}
\vspace{-4mm}
\end{table*}


\vspace{-2mm}
\noindent {\bf Tasks.} For the above two datasets, we consider the following three experimental tasks. 

\noindent {\em (i) Relationship Detection.} For ImageNet Videos, we aim at predicting a set of visual relationships with estimated subject and object trajectories. Specifically, a predicted visual relationship is counted as correct if the predicted triplet is in the ground truth set and the estimated bounding boxes have high voluminal intersection over union (vIoU) with the ground truth (vIoU threshold of $0.5$). Following~\cite{shang2017video}, this task is termed {\em relationship detection}, which contains both relationship prediction and object localization. For Charades dataset, as suggested by~\cite{sigurdsson2017asynchronous}\footnote{The performance reported in~\cite{sigurdsson2017asynchronous} refers to the mean Avergage Precision (mAP) of $157$ activities, while ours consider the detection of relation triplets. Although not being our focus, our method with the $157$ activities output achieves $33.3$ mAP on activity detection as compared to $18.3$ mAP in~\cite{sigurdsson2017asynchronous} when using only RGB frames as input. See Suppl. for details.}, we aim at detecting the visual relationships in a video without object localization, i.e.,  relationship detection happens in the scale of the entire video. For evaluation, we follow~\cite{lu2016visual,shang2017video} and adopt mean average precision (mAP) and Recall@$K$ ($K$ equals $50$ and $100$) metrics, where mAP measures the average of the maximum precisions at different recall values and Recall@$K$ measures the fraction of the positives detected in the top $K$ detection results. 

\noindent {\em (ii) Relationship Tagging.} For ImageNet Videos, the {\em relationship tagging} task~\cite{shang2017video} focuses on only relationship prediction. This is motivated by the fact that video object localization is still an open problem. Similarly, in Charades, relationship tagging focuses on only relationship prediction (where relationship tagging happens at the scale of entire video). Following~\cite{shang2017video}, we use Precision@$K$ ($K$ equals $1$, $5$, and $10$) to measure the accuracy of the tagging results. 

\noindent {\em (iii) Relationship Recognition.} Different from performing relationship reasoning at the scale of entire video, we would also like to measure how well the model recognizes the relationship in a fine-grained manner. For example, given an object trajectory and a subject trajectory, can the model predict accurate relationships? For the ImageNet Video experiments: given an input instance (with object and subject trajectories in a time segment), we measure the recognition accuracy of subject, predicate, object, and the relationship, which we term it {\em relationship recognition}. As the Charades dataset does not consider object localization, we perform recognition on object, verb, scene, and the relationship within a time segment (where relation recognition happens at the scale of a time segment in the video). We use Accuracy@$K$ ($K$ equals 1) for emphasizing whether the model gives the correct recognition result on the top 1 relationship prediction. \\




\vspace{-2mm}
\noindent {\bf Pre-Reasoning Modules} For all our experiments and ablation studies, we use the following three (exactly same) pre-reasoning modules:

\noindent {\em $\circ$ Video Chunking.} As suggested
by~\cite{carreira2017quo}, we treat the video as consecutive overlapping segments with each segment comprising continuous frames. For ImageNet Video, each segment contains $30$ frames, and adjacent segments have $15$ overlapping frames. Since the video is chunked, the object and subject trajectories are also decomposed into chunks. For Charades, each segment contains $10$ frames, and adjacent segments have $6$ overlapping frames.

\noindent {\em $\circ$ Tracklet Proposal.} Tracklet proposal is required in the ImageNet Video dataset for object localization. For each chunk in the video, we generate proposals for the possible subject and object tracklets. We utilize Faster-RCNN~\cite{girshick2015fast} as object detector trained on the $35$ objects (categories in the annotation) from MS-COCO~\cite{lin2014microsoft} and ImageNet Detection~\cite{ILSVRC15} datasets. Next, the method described in~\cite{danelljan2014accurate} is used to relate frame-level into a chunk-level object proposals. Then, non-maximum suppression (NMS) with vIoU $> 0.5$ is performed to reduce the numbers of generated chunk-level proposals. During training, proposals that have vIoU $>0.5$ with the ground truth trajectories are selected to be the training proposals. However, all the generated proposals are preserved for evaluation. 

\noindent {\em $\circ$ Feature Representation.} Following Sec.~\ref{sec:prop} notation, we express the input instance $X$ into $K$ synchronous streams of features. For the ImageNet Video, $K$ equals $3$ and the synchronous streams of features are $\{X_t^s, X_t^p, X_t^o\}_{t=1}^T$. $s, p, o$ and $T$ denote subject, predicate, object, and the number of chunks in the input instance, respectively. Note that each instance may have different numbers of chunks, i.e., different $T$, because of various duration of relationships. The output $Y_t^s, Y_t^p$ , and $Y_t^o$ follow categorical distribution. As in~\cite{shang2017video}, in the $t_{th}$ chunk of the input instance, we choose the subject and object features (i.e., $X_t^s$ and $X_t^o$) to be the averaged features for the Faster-RCNN label probability distribution outputs. $X_t^p$, on the other hand, is chosen to be the concatenation of the following three features: the improved dense trajectory (iDT) feature~\cite{wang2013action} for subject tracklet, the iDT feature for object tracklet, and the relative spatio-temporal positions~\cite{shang2017video} between subject and object tracklets. See Suppl. for more details.

For Charades, the input instance $X$ is expressed as $\{X_t^o, X_t^v, X_t^s\}_{t=1}^T$ with $o, v,$ and $s$ denoting object, verb, and scene , respectively. Since we are performing relationship reasoning directly in the entire video, we let $Y_t^o, Y_t^v$ be a multinomial distribution while $Y_t^s$ still remains to be a categorical distribution. The multinomial distribution suggests that each chunk may contain $\geq 0$ number of objects or verbs. We set $X_t^o, X_t^v,$ and $,X_t^s$ to have identical features: the output feature layer from I3D network~\cite{carreira2017quo}. See Suppl. for more details. \\


\begin{figure*}[t!]
\vspace{-4mm}
\includegraphics[width=\textwidth]{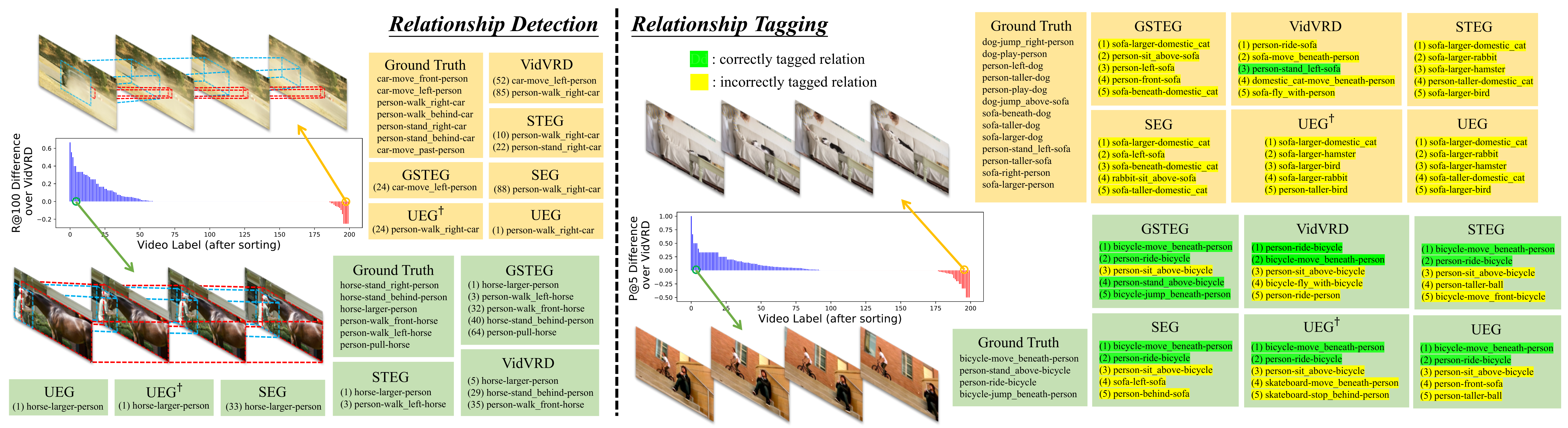}
\vspace{-5mm}
\caption{\small Examples from ImageNet Video dataset of Relationship Detection (Left) \& Tagging (Right) using baselines, ablations, and our full model. The bar plots illustrate the R@100 (left) and P@5 (right) difference comparing our model to VidVRD~\cite{shang2017video}. To show the results on all the methods, green boxes refer to a video where our model performs better and orange boxes refer to a video where VidVRD performs better. For tagging (right), we use green to highlight the correctly tagged relation and yellow for incorrectly tagged relation. The numbers in bracket represent the order of detection or tagging. Best viewed in color.}
\label{fig:qualitative}
\vspace{-4mm}
\end{figure*}


\vspace{-2mm}
\noindent {\bf Baselines} The closest baseline to our proposed model is VidVRD~\cite{shang2017video}. Beyond comparisons to~\cite{shang2017video}, we also perform a detailed ablation study of our method as well as relate to the image-based visual relationship reasoning methods (when applicable). 

\noindent {\em VidVRD.} VidVRD~\cite{shang2017video} adopted a structured loss on the multiplication of three features (i.e., $X_t^s, X_t^p$, and $X_t^o$ for ImageNet Video). The loss took softmax over all training triplets, which resembles the training objective in Visual Phrases~\cite{sadeghi2011recognition} (designed for image-based visual relationship reasoning). Note that VidVRD fails to consider the temporal structure of relationship predictions.

\noindent {\em GSTEG (Ours).} We denote our proposed method as GSTEG ({\bf G}ated {\bf S}patio-{\bf T}emporal {\bf E}nergy {\bf G}raph). For the ablation study, we choose the Energy Graph (EG) when considering different energy function designs as described below. 

\noindent {\em STEG.}  {\bf S}patio-{\bf T}emporal {\bf E}nergy {\bf G}raph (STEG) takes into account the spatial and temporal structure of video entities. However, it assumes fixed statistical dependencies between entities. Specifically, it is the non-gated version of our full model. STEG can be seen as a modified version of Asynchronous Temporal Fields (AsyncTF)~\cite{sigurdsson2017asynchronous} such that we have (1) AsyncTF's output to be a relationship prediction, and (2) 
a fully-connected spatial graph.

\noindent {\em SEG.} Compared to STEG, the {\bf S}patio {\bf E}nergy {\bf G}raph (SEG) method does not consider the temporal structure of video entities. Specifically, SEG assumes a spatially-fully-connected graph and thus the relationship predictions are made temporally independently. The counterpart in image-based visual relationship reasoning methods is Deep Relational Networks (DRN)~\cite{dai2017detecting}. We can view SEG as casting DRN to (1) take the video-based input features and (2) consider continuous object bounding boxes through time instead of a bounding box in a single frame.

\noindent {\em UEG and UEG$^\dagger$.} The {\bf U}nary {\bf E}nergy {\bf G}raph (UEG) considers the prediction of entities both spatially and temporally independently. The counterpart in image-based visual relationship reasoning methods is the Visual Relationship Detection (VRD) method of~\cite{lu2016visual} without using language priors (denoted as $VRD_V$). Similar to the modification from DRN to SEG, the accommodation from $VRD_V$ to UEG is having $VRD_V$ take the video-based features and consider object trajectories. We also perform experiments that extend UEG with additional triplet loss defined with language priors~\cite{lu2016visual}, which we denote it as UEG$^\dagger$. The counterpart in image-based methods is the full $VRD$ model of~\cite{lu2016visual}. (Please see Suppl. for more details about parameterizations and training for all the methods and datasets).

\subsection{Quantitative Analysis}
\label{sec:result}


\noindent {\em ImageNet Video.} Table.~\ref{tbl:imagenet} shows our results and comparisons to the baselines. 
We first observe that, for every metric across the three tasks (detection, tagging, and recognition), our proposed method (GSTEG) outperforms all the competing methods. 
Comparing the numbers between UEG and UEG$^\dagger$, we find that language priors can help promote visual relation reasoning. We also observe performance improvement from UEG to SEG, which could be explained by the fact that SEG  explicitly models the spatial statistical dependency in \{subject, predicate,  object\} and leads to a better relation learning between different entities. However, comparing SEG to STEG, the performance drops in some metrics, indicating that modeling temporal statistical dependency using a fixed pairwise energy parameterization may not be ideal. For example, although STEG gives a much better relationship recognition results as compared to SEG, it becomes worse in R@50 for detection and P@5 for tagging. 
This indicates that observation-gated parametrization for pairwise energy is able to capture different structure for different videos. When comparing energy graph models, VidVRD 
is able to outperform all our ablation baselines (except for the full version) in relation detection and tagging. However, it suffers from relation recognition, which requires a fine-grained understanding of visual relation in the given object and subject tracklets.


Apart from the `standard evaluation', we also considered the `zero-shot' setting, where {\em zero-shot} refers to the evaluation on the relative complement of training triplets in evaluation triplets. More specifically, in the ImageNet Video dataset, the number of all possible relation triplets is $35\times132\times35=161,700$. While the training set contains $2,961$ relation triplets (i.e., $1.83\%$ of $161,700$), the evaluation set has $1,011$ relation triplets (i.e., $0.63\%$ of $161,700$). The number of zero-shot relation triplets is $258$, which is $25.5\%$ in the evaluation set. Zero-Shot evaluation is very challenging due to the fact that we need to infer the never-seen relationship in the training set. We observe that, for most cases, our proposed method reaches the best performance compared to various baselines. The exception is mAP, where VidVRD attains the best performance using a structural objective. However, the overall trend of zero-shot evaluation mirrors standard evaluation. \\

\begin{figure*}[t!]
\vspace{-5mm}
\includegraphics[width=\textwidth]{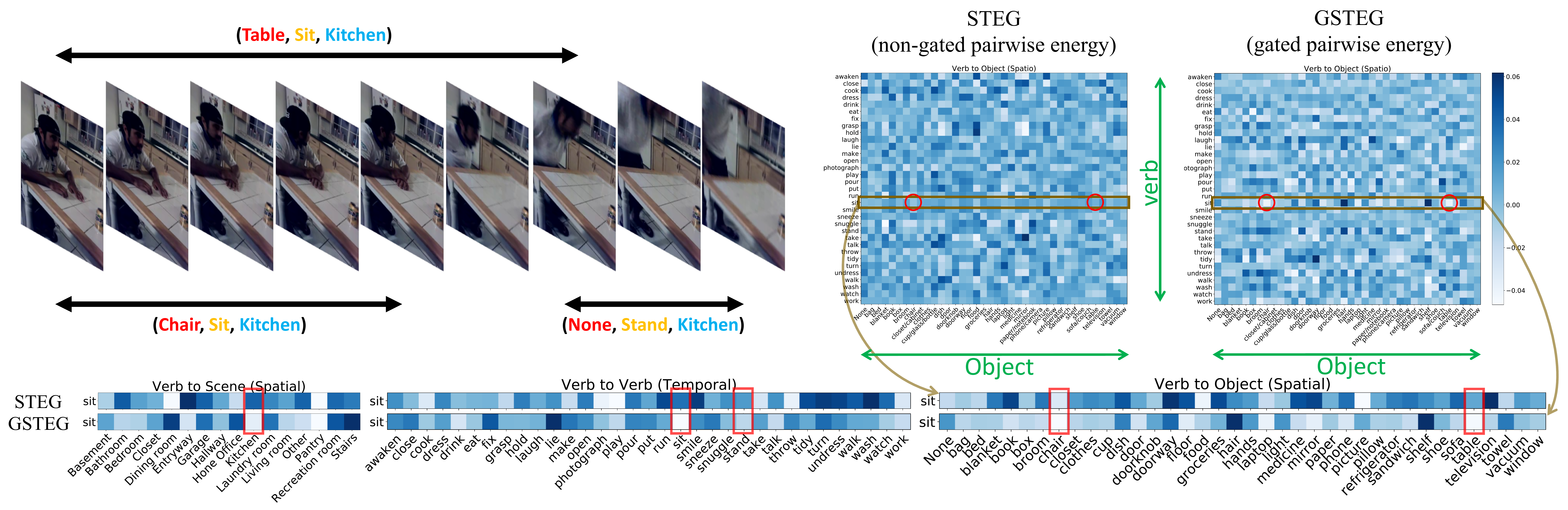}
\vspace{-5mm}
\caption{\small Analysis of non-gated and gated pairwise energies: Given an input video (top left) from Charades (that has \{{\em object, verb, scene}\} relationships), the matrices (top right) visualize the non-gated and gated pairwise energies between the verbs and objects (rows: 33 verbs, cols: 38 objects). Notice that for the verb {\em sit} (highlighted in red), the gated energy with objects {\em chair}, and {\em table} is lower compared to the corresponding non-gated pairwise energies, thereby helping towards improved relationship reasoning. A similar behavior is observed in case of verb to scene pairwise function (bottom left) as well as verb to verb pairwise function (bottom middle),  which models the temporal correlations e.g., sit/sit or sit/stand. Best viewed in color and color in the matrix or vector is normalized in its own scale.}
\label{fig:interpretation}
\vspace{-4mm}
\end{figure*}


\noindent {\em Charades.} Our results and comparisons are shown in  Table.~\ref{tbl:charades}. We find that our method outperforms all relevant baselines. We also note some interesting differences between the trend of results in Charades vs. ImageNet Video: First, comparing UEG to UEG$^\dagger$, we observe that language priors do not really help the visual relationship reasoning in Charades. We argue that it may because of the larger inter-class distinction in Charades' categories set. For example, dog/cat or horse/zebra or sit front/front/jump front share some similarity in the category set in ImageNet Video, while the categories are less semantically similar in Charades. Second, STEG constantly outperforms SEG which indicates modeling a fixed temporal statistical dependency between entities may aid the visual relationship reasoning in Charades. We hypothesize that, as compared to the ImageNet Video dataset that has a diversified set of videos in the wild between animals or inorganic substances, Charades contains videos of human indoor activities where relations between entities are much easier to model by a fixed dependency. Finally, we observe that VidVRD  performs substantially worse compared to all the other models, suggesting that the structural loss introduced by VidVRD may not generalize well to other datasets. In case of Charades, we do not perform zero-shot evaluation as the number of zero-shot relation triplets is low. \Big(The number of all the possible relation triplets is $33\times38\times16=20,064$. The training set contains $2,285$ relation triplets (i.e., $11.39\%$ of $20,064$) and the evaluation set contains $1,968$ relation triplets (i.e., $9.81\%$ of $20,064$). The number of zero-shot relation triplets is $46$, which is $2.34\%$ in the evaluation set.\Big) 

In Supplementary, we also provide the results when leveraging language priors into our model and also provide the comparisons with Structural-RNN~\cite{jain2016structural} and Graph Convolutional Network~\cite{wang2018videos}.

\vspace{-1mm}
\subsection{Qualitative Analysis}
\label{subsec:qualitative}

We next illustrate our qualitative results in Fig.~\ref{fig:qualitative} in the ImageNet Video dataset. For the relationship detection, in a scene with a person interacting with a horse, our model successfully detects $5$ out of $6$ relationships, while failing to detect horse-stand\_right-person in the top $100$ detected relationships. In another scene with a car interacting with a person, our model only detects $1$ relationship out of $7$ ground-truth relationships. We argue that the reason may be because of the sand occlusion and the small size of a person. For relationship tagging, in a scene with a person riding a bike over another person, our model successfully tags all four relationships in the top $5$ tagged results. Nevertheless, the third tagged result person-sit\_above-bicycle also looks visually plausible in this video. In another scene with a person playing with a dog on a sofa, our model fails to tag any correct relationships in the top $5$ tagged results. Our model incorrectly identified dog as cat, representing the main reason why it failed. 


Since pairwise energy in a graphical model represents the negative statistical dependency between entities, in Fig.~\ref{fig:interpretation}, for a video in Charades dataset, we provide the illustration of pairwise energy when considering our gated and non-gated parameterization. Observe that the pairwise energies between the related entities are lower for the gated parameterization as compared to the non-gated one, suggesting that the gating mechanism is able to aid video relationship reasoning by improving statistical dependency between spatially or temporally correlated entities.

\vspace{-1mm}
\section{Conclusion}

In this paper, we have presented a Gated Spatio-Temporal Energy Graph (GSTEG)
model for the task of visual relationship reasoning in videos. In the graph, we consider a spatially and temporally fully-connected structure with an amortized observation-gated parameterization for the pairwise energy functions. The gated design enables the model to detect adaptive relations between entities conditioned on the current observation (i.e., current video). On two benchmark video datasets (ImageNet Video and Charades), our method achieves state-of-the-art performance across three relationship reasoning tasks (Detection, Tagging, and Recognition).


\section*{Acknoledgement}
Work done when YHHT was in Allen Institute for AI. YHHT and RS were supported in part by the DARPA grants D17AP00001 and FA875018C0150, NSF IIS1763562, and Office of Naval Research N000141812861. LPM was  supported by the National Science Foundation (Award \# 1722822). SD and AF were supported by NSF IIS-165205,  NSF IIS-1637479, NSF IIS-1703166, Sloan Fellowship, NVIDIA Artificial Intelligence Lab, and Allen Institute for artificial intelligence. Any opinions, findings, and conclusions or recommendations expressed in this material are those of the author(s) and do not necessarily reflect the views of National Science Foundation, and no official endorsement should be inferred. We would also like to acknowledge NVIDIA’s GPU support. 

{
\small
\bibliographystyle{ieee_fullname}
\bibliography{egbib}
}

\balance

\end{document}


\title{Supplementary for Video Relationship Reasoning using \\Gated Spatio-Temporal Energy Graph}

\maketitle

\section{Towards Leveraging Language Priors}
\label{sec:transfer}

The work of~\cite{lu2016visual} has emphasized the role of language priors in alleviating the challenge of learning relationship models from limited training data. Motivated by their work, we also study the role of incorporating language priors in our framework.

In Table. 1 in the main text, comparing UEG to UEG$^\dagger$, we have seen that language priors aid in improving the relationship reasoning performance. Considering our example in Sec. 3.1 in main text, when the training instance is \{mother, pay, money\}, we may also want to infer that \{father, pay, money\} is a more likely relationship as opposed to \{cat, pay, money\} (as mother and father are semantically similar compared to mother and cat). Likewise, we can also infer \{mother, pay, check\} from the semantic similarity between {\em money} and {\em check}. 

\cite{lu2016visual} adopted a triplet loss for pairing word embeddings of object, predicate, and subject. However, their method required sampling of all the possible relationships and was also restricted to the number of entities spatially (e.g, $K=3$). Here, we present another way to make the parameterized pairwise energy also be gated by the prior knowledge in semantic space. We let the prior from semantic space be encoded as word embedding: $S = \{S^k\}_{k=1}^K$ in which $S^k \in \mathbb{R}^{|Y_t^k| \times d}$ denoting prior of labels with length $d$. We extend Eq. (3) in the main text as
\vspace{-1mm}
\begin{equation}
\footnotesize
\begin{split}
&f^\varphi_\theta (S, X_t^k,t,t',k,k',y_t^k, y_{t'}^{k'}) \\
= & f^\varphi_\theta (X_t^k,t,t',k,k',y_t^k, y_{t'}^{k'}) + u_\theta( \left \langle S^k \right \rangle_{y_t^k}) \cdot v_\theta(\left \langle S^{k'} \right \rangle_{y_{t'}^{k'}} ),
\end{split}
\label{eq:pair_prior}
\end{equation}
where $u_\theta(\cdot) \in \mathbb{R}$ and $v_\theta(\cdot) \in \mathbb{R}$ maps the label prior to a score. Eq.~\eqref{eq:pair_prior} suggests that the label transition from $Y_t^k$ to $Y_{t'}^{k'}$ can also attend to the affinity inferred from prior knowledge. 

We performed a preliminary evaluation on the relation recognition task in the ImageNet Video dataset using $300$-dim Glove features~\cite{pennington2014glove} as word embeddings. For subject, predicate, object, and relation triplet, Acc@1 metric improves from $90.60$, $28.78$, $89.79$, and $25.01$ to $90.97$, $29.54$, $90.57$, and $26.48$.

\begin{center}
\begin{figure*}[t!]
\vspace{-3mm}
  \begin{minipage}{\linewidth}
      \centering
      \begin{minipage}{0.2\linewidth}
      \vspace{-2mm}
              \includegraphics[width=\linewidth]{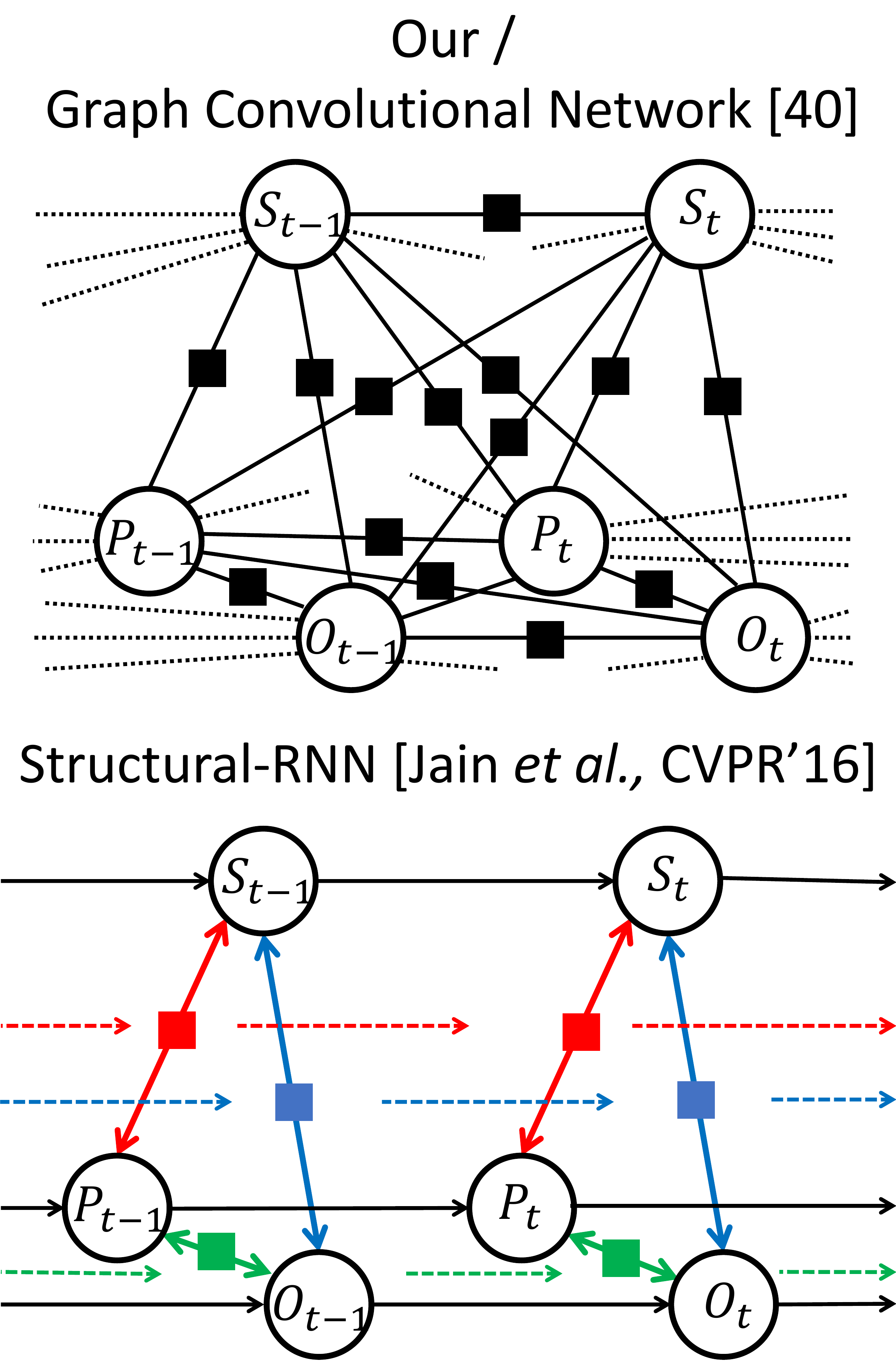}
      \end{minipage}
      \hfill \hspace{-50mm}
      \begin{minipage}{0.78\linewidth}
      \begin{center}
\fontsize{6pt}{8pt}
\selectfont
\begin{tabular}{|c||ccc||ccc||cccc|}
\hline
\multirow{3}{*}{Method} &  \multicolumn{3}{c||}{Relationship Detection}       & \multicolumn{3}{c||}{Relationship Tagging}            & \multicolumn{4}{c|}{Relationship Recognition}                           \\ &  \multicolumn{3}{c||}{relationship}         & \multicolumn{3}{c||}{relationship}            & {\em A}       &   {\em B}    & {\em C}         & relationship \\
                                &    R@50          & R@100         & mAP           & P@1            & P@5            & P@10           & Acc@1          & Acc@1          & Acc@1          & Acc@1            
                \\ \hline \hline 
\multicolumn{11}{|c|}{Standard Evaluation for ImageNet Video dataset: {\em A} = subject, {\em B} = predicate, {\em C} = object} \\ \hline \hline
Structural RNN~\cite{jain2016structural}         & 6.89 & 8.62 & 6.89 & 46.50 & 33.30 & 26.94 & 88.73 & 27.47 & 88.52 & 23.80   \\ Graph Convolution~\cite{wang2018videos}        & 6.02 & 7.53 & 8.21 & 38.50 & 30.20 & 22.70 & 86.29 & 24.22 & 85.77 & 19.18   \\ \textbf{GSTEG (Ours)}         & \textbf{7.05} & \textbf{8.67} & \textbf{9.52} & \textbf{51.50} & \textbf{39.50} & \textbf{28.23} & \textbf{90.60} & \textbf{28.78} & \textbf{89.79} & \textbf{25.01}   \\ \hline \hline 
\multicolumn{11}{|c|}{Zero-Shot Evaluation for ImageNet Video dataset: {\em A} = subject, {\em B} = predicate, {\em C} = object}   \\ \hline \hline Structural RNN~\cite{jain2016structural}          & 0.12 & 0.19 & 0.10 & 1.36 & \textbf{1.92} & 1.85 & 70.60 & 6.71 & 67.59 & 2.78   \\ Graph Convolution~\cite{wang2018videos}         & \textbf{0.16} & 0.16 & \textbf{0.20} & 1.37 & \textbf{1.92} & 1.51 & 75.00 & 5.32 & 72.45 & 3.94   \\
\textbf{GSTEG (Ours)}          & \textbf{1.16} & \textbf{2.08} & 0.15 & \textbf{2.74} & \textbf{1.92} & \textbf{1.92} & \textbf{82.18} & \textbf{7.87} & \textbf{79.40} & \textbf{6.02}   \\ \hline \hline 
\multicolumn{11}{|c|}{Standard Evaluation for Charades dataset: {\em A} = object, {\em B} = verb, {\em C} = scene}   \\ \hline \hline Structural RNN~\cite{jain2016structural}        & 23.63 & 31.15 & 8.73 & 17.18 & 12.24 & 9.18 & 42.73 & 64.32 & 34.40 & 12.40   \\ Graph Convolution~\cite{wang2018videos}         & 23.53 & 31.10 & 8.56 & 16.96 & 12.23 & 9.43 & 42.19 & \textbf{64.82} & 36.11 & 12.75   \\
\textbf{GSTEG (Ours)}          & \textbf{24.95} & \textbf{33.37} & \textbf{9.86} & \textbf{19.16} & \textbf{12.93} & \textbf{9.55} & \textbf{43.53} & \textbf{64.82} & \textbf{40.11} & \textbf{14.73}   \\ \hline
\end{tabular}
\end{center}
      \end{minipage}
  \end{minipage}
    \includegraphics[width=0.984\textwidth]{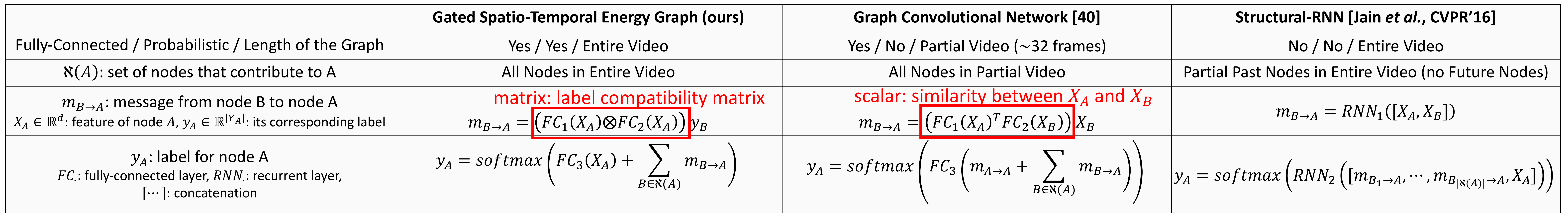}
    \vspace{0mm}
    \caption{\small [Bottom] Table summarizing the novelty of our proposed approach v.s. competing methods, [Top-left] Comparison of the graphical structures, [Top-right] Empirical comparisons between our approach and other Structural RNN~\cite{jain2016structural} and Graph Convolution~\cite{wang2018videos}. Our model performs well across all the three tasks.}
    \label{fig:comp}
\end{figure*}
\end{center}

\section{Connection to Self Attention and Non-Local Means}
\label{sec:connect}

In our main text, the message form (eq. (5)) with our observation-gated parametrization (eq. (3) with $t=t'$) can be expressed as follows:
\begin{equation*}
\footnotesize
\begin{split}
&-\mathrm{log}\,m_{t', k', t, k}(y_t^k|X) =\sum_{y_{t'}^{k'}}\varphi_{t,k,t',k'}(y_t^k, y_{t'}^{k'}|X)Q(y_{t'}^{k'})\\
\approx  & \left\{\begin{matrix}
 \sum_{y_{t'}^{k'}} \left \langle g^{kk'}_\theta(X_t^k)  \otimes h^{kk'}_\theta(X_t^k)  \right \rangle_{y_t^k, y_{t'}^{k'}} Q(y_{t'}^{k'}) & t = t' \\ \sum_{y_{t'}^{k'}} K_\sigma \Big(t,t'\Big) 
 \left \langle r^{kk'}_\theta(X_t^k)  \otimes s^{kk'}_\theta(X_t^k)\right \rangle_{y_t^k,y_{t'}^{k'}} Q(y_{t'}^{k'}) & t\neq t'
\end{matrix}\right. .
\end{split}
\end{equation*}
The equation can be reformulated in matrix form:
\begin{equation*}
 -\mathrm{log}\,{\bf m}_{t', k', t, k} 
 \approx   \mathrm{Query} \cdot \mathrm{Key}^\top \cdot \mathrm{Value},
\end{equation*}
where
\begin{equation*}
\begin{split}
& \left \langle {\bf m}_{t', k', t, k} \right \rangle_{y_t^k} = m_{t', k', t, k}(y_t^k|X) \\
& \left \langle \mathrm{Value}   \right \rangle_{y_{t'}^{k'}} = \left\{\begin{matrix}
 Q(y_{t'}^{k'}) & t = t' \\  K_\sigma \Big(t,t'\Big) \cdot Q(y_{t'}^{k'}) & t\neq t'
\end{matrix}\right. \\
& \mathrm{Query} = \left\{\begin{matrix}
 g^{kk'}_\theta(X_t^k) & t = t' \\  r^{kk'}_\theta(X_t^k) & t\neq t'
\end{matrix}\right. \\
& \mathrm{Key} = \left\{\begin{matrix}
 h^{kk'}_\theta(X_t^k) & t = t' \\  s^{kk'}_\theta(X_t^k) & t\neq t'
\end{matrix}\right. .
\end{split}
\end{equation*}

We now link this message form with self attention in Machine Translation~\cite{vaswani2017attention} and Non-Local Mean in Computer Vision~\cite{buades2005non}. Self-Attention is expressed as the form of
\begin{equation*}
\mathrm{softmax}\Big(\mathrm{Query} \cdot \mathrm{Key}^\top\Big) \cdot \mathrm{Value}
\end{equation*}
with $\mathrm{Query}$, $\mathrm{Key}$, and $\mathrm{Value}$ depending on input (termed observation in our case). 

In both Self Attention and our message form, the attended weights for $\mathrm{Value}$ is dependent on observation. The difference is that we do not have a row-wise softmax activation to make the attended weights sum to $1$. The derivation is also similar to Non-Local Means~\cite{buades2005non}. Note that Machine Translation~\cite{vaswani2017attention} focuses on the updates for features across temporal regions, Non-Local Mean~\cite{buades2005non} focuses on the updates for the features across spatial regions, while ours focuses on the updates for the entities prediction (i.e., as a message passing).

\section{Comparisons with SRNN~\cite{jain2016structural} \& GCN~\cite{wang2018videos}}

Here, we provide comparisons with Structural-RNN (SRNN)~\cite{jain2016structural} and Graph Convolutional Network (GCN)~\cite{wang2018videos} for comparisons. We note that these approaches are designed for video activity recognition, which cannot be directly applied in video visual relationship detection. In Fig.~\ref{fig:comp} (top-left), we show how we minimally modifying SRNN and GCN for evaluating them in video relationship detection. The main differences are: 1) our model constructs a fully-connected graph for entire video, while SRNN has a non-fully connected graph and GCN considers only building a graph on partial video ($\sim$32 frames), and 2) the message passing across node represents prediction's dependency for our model, while it indicates temporally evolving edge features for SRNN and similarity-reweighted features for GCN.

\section{Activity Recognition in Charades}

Sigurdsson {\em et al.}~\cite{sigurdsson2017asynchronous} proposed Asynchronous Temporal Fields (AsyncTF) for recognizing $157$ video activities. As discussed in Related Work (see Sec. 2), video activity recognition is a downstream task of visual relationship learning: in Charades, each activity (in $157$ activities) is a combination of one category  {\em object} and one category in {\em verb}. We now cast how our model be transformed into video activity recognition. First, we change the output sequence to be $Y = \{Y_t\}_{t=1}^T$, where $Y_t$ is the prediction of video activity. Then, we apply our Gated Spatio-Temporal Energy Graph on top of the sequence of activity predictions. In this design, we achieve the mAP of 33.3\%. AsyncTF reported the mAP 18.3\% for using only RGB values from a video.


\section{Feature Representation in Pre-Reasoning Modules}
\noindent {\bf ImageNet Video.} We now provide the details for representing $X_t^p$, which is the predicate feature in $t_{th}$ chunk of the input instance. Note that we use the relation feature from prior work~\cite{shang2017video} (the feature can be downloaded form~\cite{VidVRDURL}) as our predicate feature. The feature comprises three components: (1) improved dense trajectory (iDT) features from subject trajectory, (2) improved dense trajectory (iDT) features from object trajectory, and (3) relative features describing the relative positions, size, and motion between subject and object trajectories. iDT features are able to capture the movement and also the low-level visual characteristics for an object moving in a short clip. The relative features are able to represent the relative spatio-temporal differences between subject trajectory and object trajectory. Next, the features are post-processed as bag-of-words features after applying a dictionary learning on the original features. Last, three sub-features are concatenated together for representing our predicate feature.

\noindent {\bf Charades.} We use the output feature layer from I3D network~\cite{carreira2017quo} to represent our object ($X_t^o$), verb ($X_t^v$), and scene feature ($X_t^s$). The I3D network is pre-trained from Kinetics dataset~\cite{carreira2017quo} (the model can be downloaded from~\cite{I3DURL}) and the output feature layer is the layer before output logits.
 
\section{Intractable Inference during Evaluation}
In ImageNet Video dataset, during evaluation, for relation detection and tagging, we have to enumerate all the possible associations of subject or object tracklets. The number of possible associations grows exponentially by the factor of the number of chunks in a video, which will easily become computationally intractable. Note that the problem exists only during evaluation since the ground truth associations (for subject and object tracklets) are given during training. To overcome the issue, we apply the greedy association algorithm described in~\cite{shang2017video} for efficiently associating subject or object tracklets. The idea is as follows. First, we adopt the inference only in a chunk. Since the message does not pass across chunks, at this step, we don't need to consider associations (for subject or object tracklets) across chunks. In a chunk, for a pair of subject and object tracklet, we have a predicted relation triplet. Then, from two overlapping chunks, we only associate the pair of the subject and object tracklets with the same predicted relation triplet and high tracklets vIoU (i.e., $>0.5$). Comparing to the original inference, this algorithm exponentially accelerates the time computational complexity. On the other hand, in Charades, we do not need associate object tracklets. Thus, the intractable computation complexity issue does not exist. The greedy associate algorithm is not required for Charades.

\section{Training and Parametrization Details}
We specify the training and parametrization details as follows.

\noindent {\bf ImageNet Video.} Throughout all the experiments, we choose Adam~\cite{kingma2014adam} with learning rate $0.001$ as our optimizer, $32$ as our batch size, $30$ as the number of training epoch, and $3$ as the number of message passing. We initialize the marginals to be the marginals estimated from unary energy.
\begin{itemize}
    \item Rank number $r$: 5
    \item $g_\theta^{kk'}(X_t^k)$: $|X_t^k|\times (|Y_t^k| \times r)$ fully-connected layer, resize to  $|Y_t^k| \times r$
    \item $h_\theta^{kk'}(X_t^k)$: $|X_t^k|\times \times (|Y_{t'}^{k'}| \times r)$ fully-connected layer, resize to  $|Y_{t'}^{k'}| \times r$
    \item $r_\theta^{kk'}(X_t^k)$: $|X_t^k|\times 1024$ fully-connected layer, ReLU Activation, Dropout with rate $0.3$, $1024 \times 1024$ fully-connected layer, ReLU Activation, Dropout with rate $0.3$, $1024 \times (|Y_t^k| \times r)$ fully-connected layer, resize to  $|Y_t^k| \times r$
    \item $s_\theta^{kk'}(X_t^k)$: $|X_t^k|\times 1024$ fully-connected layer, ReLU Activation, Dropout with rate $0.3$, $1024 \times 1024$ fully-connected layer, ReLU Activation, Dropout with rate $0.3$, $1024 \times (|Y_{t'}^{k'}| \times r)$ fully-connected layer, resize to  $|Y_{t'}^{k'}| \times r$
    \item $\sigma$: $10$
    \item $w_\theta^{kk'}(X_t^k)$: $|X_t^k|\times|Y_t^k|$ fully-connected layer 
\end{itemize}

\noindent {\bf Charades:} Throughout all the experiments, we choose SGD with learning rate $0.005$ as our optimizer, $40$ as our batch size, $5$ as the number of training epoch, and $5$ as the number of message passing. We initialize the marginals to be the marginals estimated from unary energy.
\begin{itemize}
    \item Rank number $r$: 5
    \item $g_\theta^{kk'}(X_t^k)$: $|X_t^k|\times (|Y_t^k| \times r)$ fully-connected layer, resize to  $|Y_t^k| \times r$
    \item $h_\theta^{kk'}(X_t^k)$: $|X_t^k|\times  (|Y_{t'}^{k'}| \times r)$ fully-connected layer, resize to  $|Y_{t'}^{k'}| \times r$
    \item $r_\theta^{kk'}(X_t^k)$: $|X_t^k|\times (|Y_t^k| \times r)$ fully-connected layer, resize to  $|Y_t^k| \times r$
    \item $s_\theta^{kk'}(X_t^k)$: $|X_t^k|\times \times (|Y_{t'}^{k'}| \times r)$ fully-connected layer, resize to  $|Y_{t'}^{k'}| \times r$
    \item $\sigma$: $300$
    \item $w_\theta^{kk'}(X_t^k)$: $|X_t^k|\times|Y_t^k|$ fully-connected layer 
\end{itemize}

\section{Parametrization in Leveraging Language Priors}
Additional networks in the experiments towards leveraging language priors are parametrized as follows:
\begin{itemize}
    \item $d$: 300 (because we use 300-dim. Glove~\cite{pennington2014glove} features)
    \item $u_\theta (\cdot)$:  $d\times 1024$ fully-connected layer, ReLU Activation, Dropout with rate $0.3$, $1024 \times 1024$ fully-connected layer, ReLU Activation, Dropout with rate $0.3$, $1024 \times 1$ fully-connected layer
    \item $v_\theta (\cdot)$:  $d\times 1024$ fully-connected layer, ReLU Activation, Dropout with rate $0.3$, $1024 \times 1024$ fully-connected layer, ReLU Activation, Dropout with rate $0.3$, $1024 \times 1$ fully-connected layer
\end{itemize}

\section{Category Set in Dataset}
\label{sec:categ}

For clarity, we use bullet points for referring to the category choice in datasets for the different entity.

\begin{itemize}
    \item {\em subject} / {\em object} in ImageNet Video (total $35$ categories)
        \begin{itemize}
            \item airplane, antelope, ball, bear, bicycle, bird, bus, car, cat, cattle, dog, elephant, fox, frisbee, giant panda, hamster, horse, lion, lizard, monkey, motorcycle, person, rabbit, red panda, sheep, skateboard, snake, sofa, squirrel, tiger, train, turtle, watercraft, whale, zebra
        \end{itemize}
    \item {\em predicate} in ImageNet Video (total $132$ categories)
        \begin{itemize}
            \item taller, swim behind, walk away, fly behind, creep behind, lie with, move left, stand next to, touch, follow, move away, lie next to, walk with, move next to, creep above, stand above, fall off, run with, swim front, walk next to, kick, stand left, creep right, sit above, watch, swim with, fly away, creep beneath, front, run past, jump right, fly toward, stop beneath, stand inside, creep left, run next to, beneath, stop left, right, jump front, jump beneath, past, jump toward, sit front, sit inside, walk beneath, run away, stop right, run above, walk right, away, move right, fly right, behind, sit right, above, run front, run toward, jump past, stand with, sit left, jump above, move with, swim beneath, stand behind, larger, walk past, stop front, run right, creep away, move toward, feed, run left, lie beneath, fly front, walk behind, stand beneath, fly above, bite, fly next to, stop next to, fight, walk above, jump behind, fly with, sit beneath, sit next to, jump next to, run behind, move behind, swim right, swim next to, hold, move past, pull, stand front, walk left, lie above, ride, next to, move beneath, lie behind, toward, jump left, stop above, creep toward, lie left, fly left, stop with, walk toward, stand right, chase, creep next to, fly past, move front, run beneath, creep front, creep past, play, lie inside, stop behind, move above, sit behind, faster, lie right, walk front, drive, swim left, jump away, jump with, lie front, left
        \end{itemize}
    \item {\em verb} in Charades (total $33$ categories)
        \begin{itemize}
            \item awaken, close, cook, dress, drink, eat, fix, grasp, hold, laugh, lie, make, open, photograph, play, pour, put, run, sit, smile, sneeze, snuggle, stand, take, talk, throw, tidy, turn, undress, walk, wash, watch, work
        \end{itemize}
    \item {\em object} in Charades (total $38$ categories)
        \begin{itemize}
            \item None, bag, bed, blanket, book, box, broom, chair, closet/cabinet, clothes, cup/glass/bottle, dish, door, doorknob, doorway, floor, food, groceries, hair, hands, laptop, light, medicine, mirror, paper/notebook, phone/camera, picture, pillow, refrigerator, sandwich, shelf, shoe, sofa/couch, table, television, towel, vacuum, window
        \end{itemize}
    \item {\em scene} in Charades (total $16$ categories)
        \begin{itemize}
            \item Basement, Bathroom, Bedroom, Closet / Walk-in closet / Spear closet, Dining room, Entryway, Garage, Hallway, Home Office / Study, Kitchen, Laundry room, Living room, Other, Pantry, Recreation room / Man cave, Stairs
        \end{itemize}
\end{itemize}

{
\small
\bibliographystyle{ieee_fullname}
\bibliography{egbib}
}